\documentclass{article}

\usepackage{PRIMEarxiv}

\usepackage[utf8]{inputenc} % allow utf-8 input
\usepackage[T1]{fontenc}    % use 8-bit T1 fonts
\usepackage{hyperref}       % hyperlinks
\usepackage{url}            % simple URL typesetting
\usepackage{booktabs}       % professional-quality tables
\usepackage{amsfonts}       % blackboard math symbols
\usepackage{nicefrac}       % compact symbols for 1/2, etc.
\usepackage{microtype}      % microtypography
\usepackage{lipsum}
\usepackage{fancyhdr}       % header
\usepackage{graphicx}       % graphics
\graphicspath{{media/}}     % organize your images and other figures under media/ folder
\usepackage{amsmath}

\title{Meta Semantics: Towards better natural language understanding and reasoning
}

\author{
 Xiaolin Hu \\
  Institute of Computing and Mathematical Sciences \\
  University of Leicester \\
  Leicester\\
  \texttt{xh152@le.ac.uk} \\
 }

\begin{document}
\maketitle

\begin{abstract}
Natural language understanding is one of the most challenging topics in artificial intelligence. Deep neural network methods, particularly large language module (LLM) methods such as ChatGPT and GPT-3, have powerful flexibility to adopt informal text but are weak on logical deduction and suffer from the out-of-vocabulary (OOV) problem. On the other hand, rule-based methods such as Mathematica, Semantic web, and Lean, are excellent in reasoning but cannot handle the complex and changeable informal text. Inspired by pragmatics and structuralism, we propose two strategies to solve the OOV problem and a semantic model for better natural language understanding and reasoning.
\end{abstract}

% keywords can be removed
\keywords{Natural language understanding \and automated reasoning \and word embedding}

\section{Introduction}
Natural language understanding is the study of making machines understand the daily used informal text. There are two main categories of methods, statistic-based methods and rule-based methods. Benefiting from the blow-up of deep learning algorithms such as transformer\cite{vaswani2017attention}, the statistic-based methods upgrade from the traditional Bayesian methods and have better robustness. On the hand, the rule-based methods are wildly used in expert systems, which are run by handwritten rules from experts and use the patterns to map the natural language to machine-readable commands such as SQL, the LUNAR system\cite{woods1973progress}, as an example, which is used in the analysis of lunar geology. Although both methods have got great achievements, there still exist some main challenges that we need to resolve. In section 2, we will discuss the success and challenges of the existing natural language understanding models. In section 3, a potential solution to the OOV problem from word embedding which limits the deep neural method to reasoning and understanding will be presented. In section 4, we will propose our semantic model in detail to move the natural language understanding into the next stage. In section 5, we will expound on the rationality of the Meta Semantic model from the perspective of pragmatics and epistemology.

\section{Related work}

\subsection{Statistic-based methods}

GPT-3\cite{brown2020language} is a large natural language model based on the transformer which is pre-trained on trillions text data with 175 billion parameters and has been fine-tuned for several downstream tasks, such as translation, sentiment analysis, and story writing. Then based on GPT-3, openAI presents several derivative versions such as Codex and ChatGPT which are specialized in code generation and user interaction. GPT-series shows amazing performance in daily communication and text generation such as stories, news and even generating a workable programming code. The success of GPT heralds a new era of natural language processing with LLM as the core. However, most texts generated by GPT are only “convincing” rather than “true”. Statistical models always seem to be very weak when dealing with mathematics-related problems\cite{frieder2023mathematical}, even digital-related issues. The main issue causing this phenomenon is the OOV problem we discuss in the later section.

\subsection{Rule-based methods}

The rule-based methods focus on the knowledge behind the text. They use formal text to express the information and reason for new information. Mathematica\cite{wolfram1991mathematica} and Isabelle\cite{nipkow2002isabelle} used functional programming based on lambda calculus. These rule-based systems are wildly used in symbolic systems and theorem provers, benefiting from the safe and stable feature of functional programming. Semantic web \cite{berners2001semantic} and knowledge graphs\cite{chen2020review}, use tuples to represent the relation between the entities, by searching on the graph path. In this way, the new knowledge is deducted. However, the rule-based systems need a number of man-craft rules or relation tuples, which are time-consuming and hard to maintain. Moreover, in the theorem proving systems such as lean\cite{de2015lean}, Isabelle and Coq\cite{bertot2013interactive}, the input and output are not in natural language, which limits the broad application of those powerful systems.

\subsection{Auto-formalization}

Automatic formalization takes on the task of mapping natural language to formal context. At the moment, the primary research topics in automatic formalization are Math Word Problems\cite{kushman2014learning,wang2018translating}, Code Generation\cite{chen2021evaluating}, and natural language to predicate logic\cite{levkovskyi2021generating, liang2016learning} or first-order logic\cite{lu2022parsing}. However, the following are the primary issues that automatic formalization faces: 1. Inadequate training data 2. An excessive emphasis on end-to-end and a lack of excessive processes in the middle (analysis). In the Math Word Problem, for example, a training data from dataset Alg514 is as Table 1:\\
\begin{table}[h]
\centering
\begin{tabular}{|l|l|}
\hline
Question  & \begin{tabular}[c]{@{}l@{}}An employment agency specializing in temporary construction help pays heavy equipment\\ operators 140 dollars per day and general laborers 90 dollars per day. If 35 people\\ were hired and the payroll was 3950 dollars, how many heavy equipment operators were \\ employed? How many laborers were employed?\end{tabular} \\ \hline
Equations & \begin{tabular}[c]{@{}l@{}}140 * operators + 90 * laborers = 3950\\ laborers + operators = 35\end{tabular} \\ \hline
Solutions & 16, 19 \\ \hline
\end{tabular}
\caption{Data from Alg514 for Math Word Problem}
\end{table}
\\
There is a significant lack of appropriate intermediate process statements from the problem to the equations and solution, such as "Heavy equipment is paid 140 dollars per day per person," "The number of people must be an integer," and so on. Despite the fact that large pre-trained language models address the lack of intermediate processes by taking general knowledge. However, because its learning content has not been formalized or understood in a formal manner, the content it represents only has a formal appearance and no actual understanding or application\cite{frieder2023mathematical}.

\section{Solution to the OOV problem}
The out-of-vocabulary (OOV) problem in deep neural methods arises mainly for the following two reasons\cite{jiang2019learning}:
\begin{itemize}
    \item Numbers are infinite and are the main source of OOV in the dataset, and deep language models rely on softmax to predict the generated words. However, softmax can only make predictions within a finite vocabulary, so numbers cannot be predicted as a whole or individual token.
    \item The current number reading method is to split the number and read it one by one. Because there is no context relationship between the various bits of the number, there is an inaccuracy of number generation due to the loss of information in the statistic-based generation method.
\end{itemize}
 To solve the problem, we use three assumptions.\par

1. Most of the OOV must be a deduction of an existing word or will be defined by the later semantic.\par
2. The semantic or embedding of OOV is equal to the abstraction of the deduction content.\par
3. There exist some OOV only take the role as formal parameters.
The OOV appearing in the text can be roughly divided into the following two cases to be solved separately: OOV comes from rule deduction and OOV comes from the input directly.

\subsection{OOV comes from rule deduction}

There are numerous OOV in the context generated by rule application. In a symbolic system, the result of "1 + 1" is unique. After compiling and calculating 2 is the unique answer. However, in other contexts, the meaning of “1 + 1” is just a writing of the formula “1 + 1”. Therefore, the functions of “+” are abundant. Suppose there are at least two individual concepts of plus, the computational plus and writing plus. Assume “2” is an OOV, “1”, and “computational plus” are in-dictionary vocabulary (IDV). In this case, the OOV “2” could be deducted by a rule-based system using the function of "computational plus". Therefore, the OOV “2” is the deduction of tuple "<1, computational plus, 1>"\par

\begin{equation}
   <1, computational \ plus, 1> \to 2
   \nonumber 
\end{equation}

Let $v(i)$ as any word embedding in dictionary vocabulary. $v(o)$ as the word embedding of out of vocabulary. Since the OOV could be deduced by limited IDV. We assume the semantic or word embedding of the deduction equals the semantic of the abstraction of the original tuple.

\begin{equation}
\begin{split}
    <i_1, i_2 , i_3> \to o\\
    v(o) = abstract(v(i_1), v(i_2), v(i_3))
    \nonumber 
\end{split}
\end{equation}

In this case, it is:

\begin{equation}
    v("2") = abstract(v("1"), v("computational \ plus"), v("1"))
    \nonumber 
\end{equation}

And same as the OOV formula “1 + 1”, the embedding or semantic could be represented by 

\begin{equation}
    v("1 + 1") = abstract(v("1"), v("writing \ plus"), v("1"))
    \nonumber 
\end{equation}

Need to notice that the dimension of $v(o)$ is equal to the dimension of $v(i)$. $ shape(v(o)) = shape(v(i)) = 1 \times D $, where $ D $ is the embedding dimension. In this way, all the embedding of OOV that could be deducted from the rule-based system could be represented.\par

\subsection{OOV comes from the input directly}

Some OOV is obtained from the input directly, which has no inference in the preceding text. For example, "We have 211, it is a prime number" Assume "211" is the OOV that was obtained directly from the user rather than through system deduction. Lambda substitution is used to solve this problem. \par

Assume there are n lambda words for substitution in the dictionary, $lambda_1, ..., lambda_n$. When the system receives an OOV that is not derived from the system, it will be assigned to a $lambda i$. As a result, the input is rewritten as

\begin{center}
    “We have $lambda_i$, it is a prime number”
\end{center}

When $lambda_i$ is generated, it is directly substituted to the assigned content. "211" in this case. To keep the unassigned lambda from appearing, a filter is added to remove the empty lambda. \par

\begin{equation}
    p(lambda_i = NULL | w_1 ... w_n) = 0
    \nonumber 
\end{equation}

To avoid making the dictionary too large and keep it under control. The deduction process can be stored in an external knowledge base for those low-frequency, deductible concepts. When lambda accepts an OOV, the system may search the knowledge base for this OOV to see if there is a deduction process for this concept, or the system may use an external rule-based method to decompose the OOV into a combination of IDV. The OOV problem in word embedding is solved in this way by using a hybrid scheme of rule-based and statistical methods.

\section{Meta Semantics}

To have the AI understand things in logic, it is preferable to make it grasp things in a formal way, which makes reasoning easier and allows the process of deduction to be observable and human-readable. To realise the aforementioned aim, we first propose the Meta Semantic model.

\subsection{Grammar}
In this semantic model, meta is used to represent every fact (understanding). Every meta includes three parts: subject, predicate, and object. And each part could be a meta or concept. The structure of meta follows the context-dependency grammar. 

\begin{equation}
\begin{split}
        Meta \to (SUBJ, PRED, OBJ)\\
        SUBJ, PRED, OBJ \to Meta, Concept
        \nonumber 
\end{split}
\end{equation}
For any concept, it represents a rule-based function. Which has the ability to deduct or abstract the semantics of the meta that take this concept as a predicate.

\begin{equation}
    PRED(SUBJ, OBJ) \to DEDUCTION
    \nonumber 
\end{equation}

By this property, Meta Semantic could do inference using concept corresponding function. Like
\begin{equation}
    ("1", "computational plus", "1") \to "2"
    \nonumber
\end{equation}

The new concept “2” takes the semantic of abstract ("1", "computational plus", "1"). For that meta that could not be deducted by rules, the system just combines the content of each concept. Such as
\begin{equation}
    ("Jack", "love", "Rose") \to "Jack \ love \ rose".
    \nonumber 
\end{equation}

And new abstract concept “Jack love rose” takes the semantic of abstract ("Jack", "love", "rose"). The above reasoning is named rule-based reasoning. Because not all reasoning is deduced by the rule-based method, most reasoning is not rule-based, which is proposed as empirical reasoning. For instance:

\begin{equation}
        (((All, det, man), (will, decorate, be), die), and, (Socrates, be, man)) \dashrightarrow (Socrates, (will, decorate, be), die)
    \nonumber 
\end{equation}
In order for the model to automatically derive the meta that the rule-based system can calculate. The system needs to learn formal statements in order to derive and use subsequent derivations. As an example:

\begin{equation}
    (calculate, apply on, (lambda_1, +, lambda_2)) \dashrightarrow ((lambda_1, computational plus, lambda_2), \to, lambda_3)
    \nonumber
\end{equation}
where $\dashrightarrow$ indicates the deduction is an empirical deduction and $\to$ indicates a rule-based deduction. In this way, the deduction result of rule-based induction is assigned to $lambda_3$.

\subsection{Semantic Embedding}

The concept is embedded by the Meta Semantic Model in two steps: base embedding and general embedding. The goal of base embedding is to learn the relationship at its most fundamental level. The general embedding refines the base embedding into a general embedding that can accept sentence-level relations and abstract the meta into an embedding. \par

\subsubsection{Base Embedding}
The system begins with the base meta statement in training data before moving on to high-level abstraction semantics. If a meta is created solely by a concept. This is referred to as base meta. Assume the dictionary contains N concepts; each concept could be represented by a one-hot code with a dimension of $1 \times N$. The CBOW\cite{goldberg2014word2vec} model is then used to train the IDV's base embedding, which displays the co-occurrence and position information of the concept in the base meta. \par
\begin{figure}[h]
  \centering
  \includegraphics[scale=0.5]{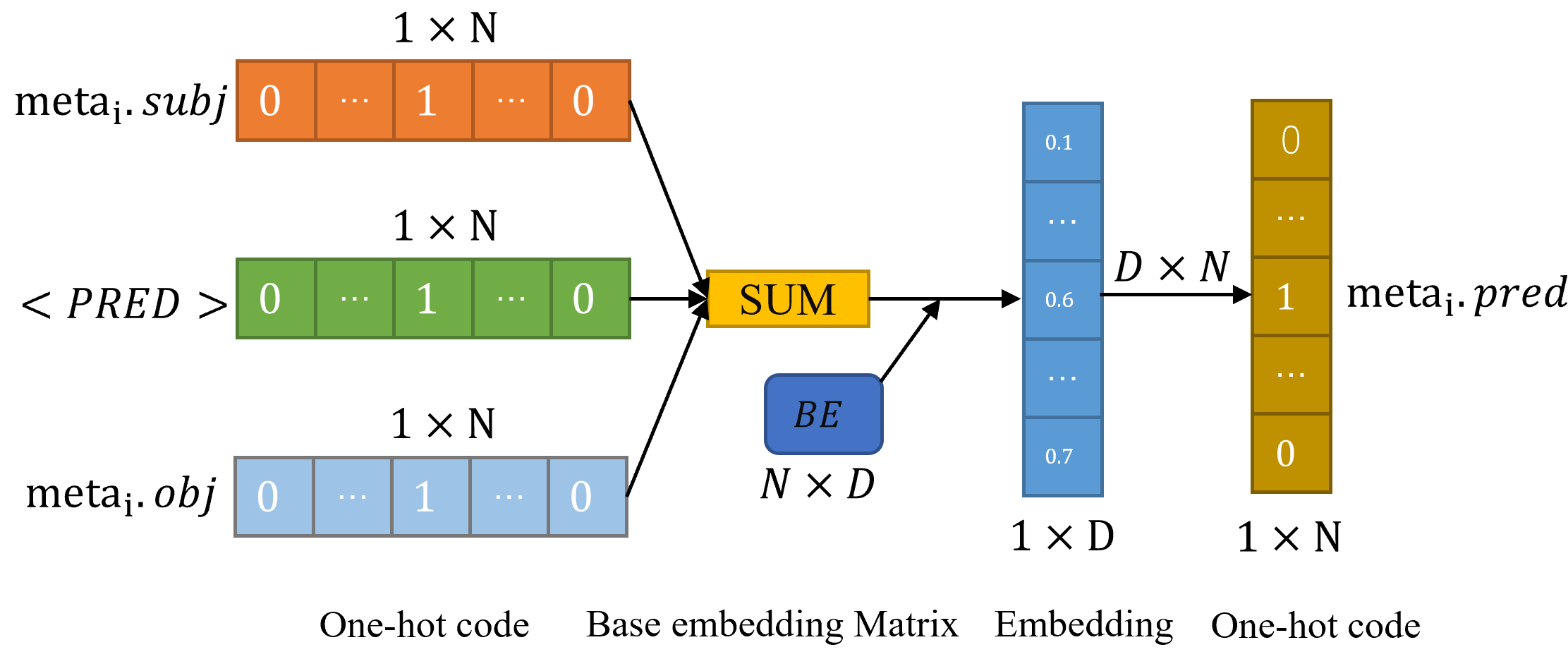}
  \caption{Base Embedding}
\end{figure}

\subsubsection{General Embedding}
The system then uses the concept's base embedding as the target to train the general embedding matrix. The general embedding matrix performs two functions: abstracting the meta semantics and embedding the concept at the sentence level.
\begin{figure}[h]
  \centering
  \includegraphics[scale=0.5]{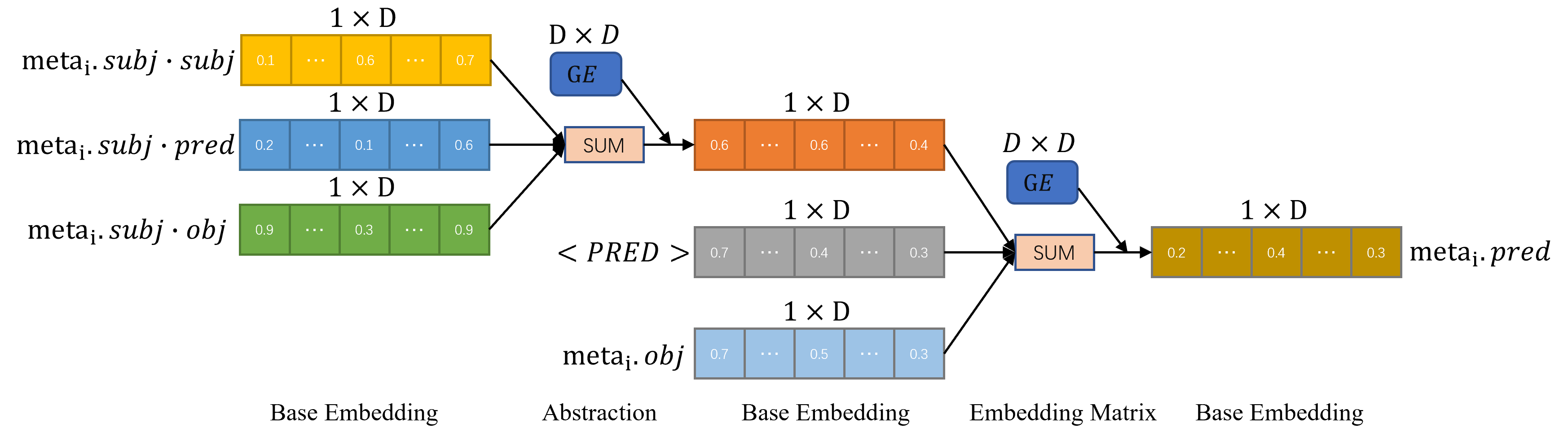}
  \caption{Abstraction and Embedding}
\end{figure}

\subsection{Training}
Meta Semantic learns embedding unsupervised by breaking down a base meta into 18 small tasks as Table 2.

\begin{table}[h]
\centering
\begin{tabular}{l}
\hline
(<SUBJ>, pred, obj) $\dashrightarrow$ subj\\
(subj, <PRED>, obj) $\dashrightarrow$ pred\\
(subj, pred, <OBJ>) $\dashrightarrow$ obj\\
(<SUBJ>, <SUBJ-PRED>, pred) $\dashrightarrow$ subj\\
(<SUBJ>, <SUBJ-OBJ>, obj) $\dashrightarrow$ subj\\
(<PRED>, <PRED-SUBJ>, subj) $\dashrightarrow$ pred\\
(<PRED>, <PRED-OBJ>, obj) $\dashrightarrow$ pred\\
(<OBJ>, <OBJ-SUBJ>, subj) $\dashrightarrow$ obj\\
(<OBJ>, <OBJ-PRED>, pred) $\dashrightarrow$ obj\\
(subj,<RELATION>, subj) $\dashrightarrow$ <ID>\\
(subj, <RELATION>, pred) $\dashrightarrow$ <SUBJ-PRED>\\
(subj, <RELATION>, obj) $\dashrightarrow$ <SUBJ-OBJ>\\
(pred, <RELATION>, subj) $\dashrightarrow$ <PRED-SUBJ>\\
(pred, <RELATION>, pred) $\dashrightarrow$ <ID>\\
(pred, <RELATION>, obj) $\dashrightarrow$ <PRED-OBJ>\\
(obj, <RELATION>, subj) $\dashrightarrow$ <OBJ-SUBJ>\\
(obj, <RELATION>, pred) $\dashrightarrow$ <OBJ-PRED>\\
(obj, <RELATION>, pred) $\dashrightarrow$ <ID>\\ \hline
\end{tabular}
\caption{18 derivative tasks from base meta}
\end{table}

The base embedding is trained as a classification task to get a better-qualified base embedding.
\begin{equation}
    L=\ -\log{p(w_t|w_{i_1},w_{i_2},w_{i_3})}
\end{equation}
Where $w_t$ is the one hot code of the deduction concept and $w_{i_n}$ is the one hot code of each part of the input meta. To improve flexibility and inclusiveness in general embedding, the training of general embedding is a regression task.
\begin{equation}
    L=\ \frac{1}{D}\sum_{k=1}^{D}\left|{v_t}_k-{u_t}_k\right|^2
\end{equation}
where $v_{t_k}$ is the $k$ row of the base embedding of deduction concept $t$. $u_{t_k}$ is the $k$ row of the embedding prediction of concept $t$.

\subsection{Semantic Decoding}
The Meta Semantic generation can be considered as a tree2tree generation model. The generation of the tree can be regarded as the decoding of the following embedding.
\begin{equation}
\begin{split}
    v(output) = abstract(v(input))\\
    output = Decode(v(output))
    \nonumber
\end{split}
\end{equation}

For any embedding, $v(input)= 1 \times D$. There is a terminal discriminator that determines whether the semantic tread is a concept or a Meta. If it's a concept, find the one with the most similar semantics. If it is a meta. Decode the embedding of meta to subj, pred, and obj and keep decoding those outputs until it ends with a concept using three decoders. As a result, the system could generate a potential explanation for a specific semantic.

\begin{figure}
  \centering
  \includegraphics[scale=0.5]{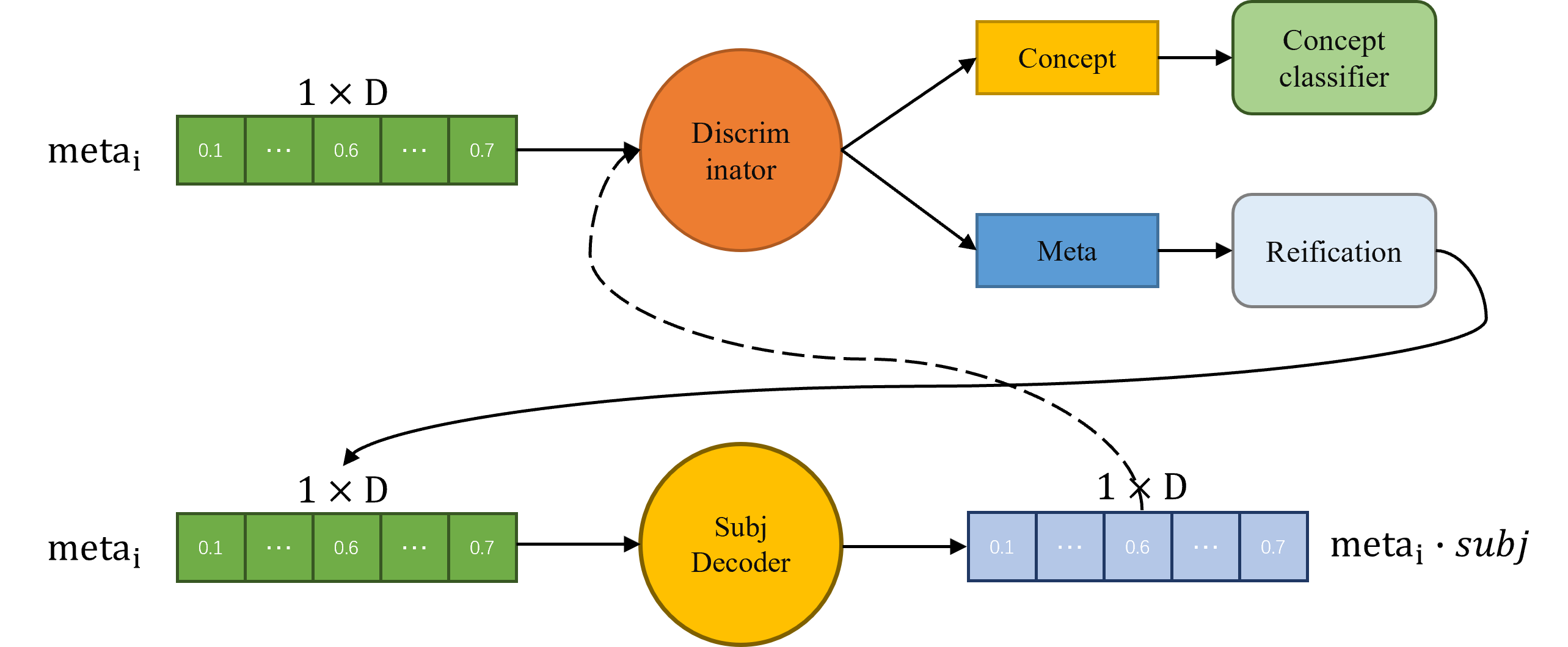}
  \caption{Decoder}
\end{figure}

\section{Reflections on Human Cognition and AI}
Artificial intelligence is more than a mathematical and computer science problem. The development of philosophy and linguistics cannot be separated from the guiding ideology of the design of AI. From early Aristotle's formal proof of syllogism, which laid the foundation for formal logic, to postmodern Ferleger predicate calculus, Russell's logical atomism, and Chomsky's formal language theory, logic has accompanied the formalized cognitive process. This section focuses on the epistemology of the natural language understanding model and cognition. The following three points will be discussed in depth:
\begin{quote}
1. Semantics and cognition have more complete information than language.\par
2. Cognition can be formalized under the premise that cognition is discretely conceptualized and the cognition process has a limited window.\par
3. The basic understanding and process of humans are some formalized small tasks and small logic while generating reasoning through experience.
\end{quote}

\subsection{Pragmatics view on machine understanding}

The natural language understanding research object is not only plain text but also the restoration of the cognitive process through language. Pragmatics asserts that natural language cannot be understood literally, and that context and speech act objects make significant differences in understanding language semantics\cite{jurafsky2004pragmatics, levinson1983pragmatics}. As a result, pragmatics has three main research objects: language, context, and speech act objects. Language can be considered a static object in a short period of time, whereas context and speech act objects are more complex. Context can range from the start and end of a conversation to the entire life experience of the speech act object. Similarly, speech act objects differ in a variety of ways. The essence of natural language understanding is to restore the cognitive process of the speech act object itself. However, it is impossible to model and analyze all individuals and which will also make the result of understanding personal. Therefore, the research object of natural language understanding should be an ideal individual, whose context is all the training samples, and whose understanding should be a widely recognized social consensus understanding, which can be said to be a rational understanding to a certain extent.

\subsection{Discrete cognition leads to conceptualized and formal cognition}

Humans can only cognize discretely and think in discrete steps. Obviously, this statement does not appear to be true; time, displacement, continuous function, and all other forms of continuous existence that humans are capable of comprehending strongly contradict this statement. However, what this sentence really means is that if humans want to understand a being, they must discretize it or treat it as a distinct individual. When something is referred to as "a thing," it means it already exists discretely, is recognised individually, or has been conceptualised. To put it another way, we understand the world by recognising discrete, individual concepts.\par

For example, it is known that line is a continuous existence. However, when people see a line, they simply regard it as an instance of concept—line, rather than the infinity combinations of points, and it made no difference how different the line was from any other line he had seen before in terms of treating it as an individual concept. This abstraction and conceptualization process allows humans to stop thinking about the infinite details and instead see existence as a bounded, internal continuum. In short, one realizes continuity so that one can only think about “it” instead of thinking about the infinite of “them”.\par
Other aspects can also support this discrete cognition. The output of VR devices, for example, are discrete frames, but the human brain can perceive the images as continuous. To some extent, this demonstrates that human perception is limited and that this limitation is also the cause of discrete and conceptualized cognition. The clue of conceptualization can also be found in the image segment. The success of the unsupervised image segment shows that concepts have their nature feature to be discrete into individual concepts as Figure 4 shows. \par

\begin{figure}
  \centering
  \includegraphics[scale=0.23]{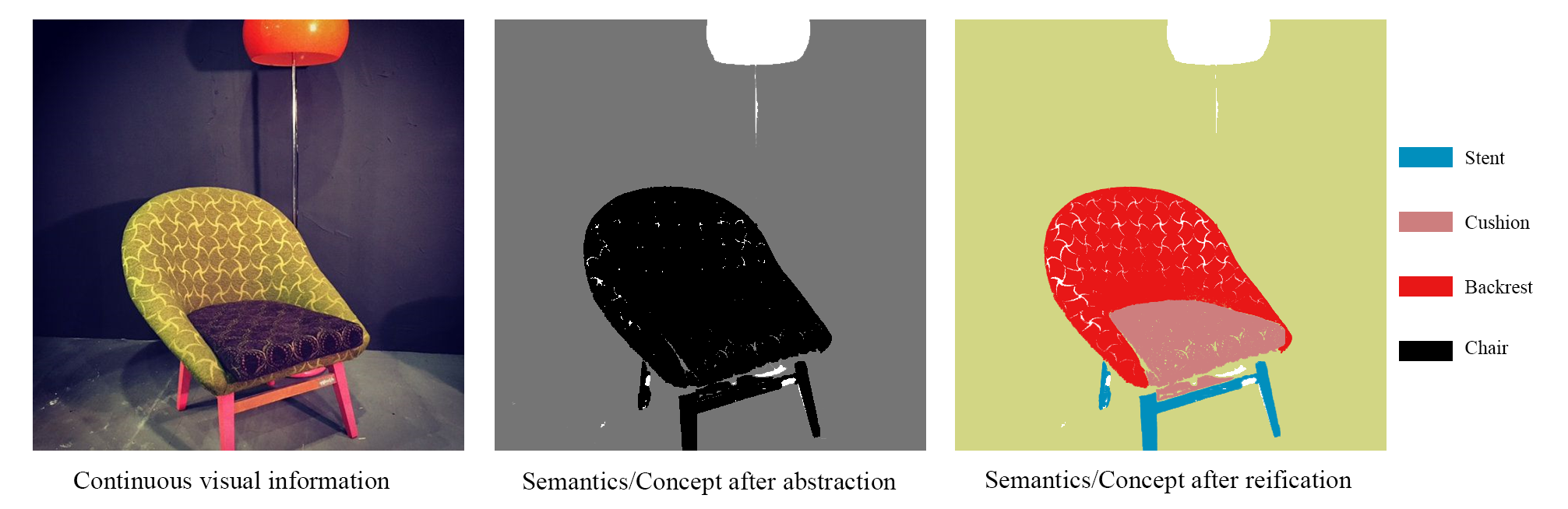}
  \caption{The conceptualization in image segment}
\end{figure}

Once conceptualized cognition is recognised, human perception can be viewed as a series or multiple parallel discrete sequences of concepts, and human cognition is formalised to some extent. It is important to note that conceptualization does not require identity to name a perception or find a corresponding concept for it; as long as it can see a thing as a whole, the whole is a concept. \par
Based on the assumption that cognitive sequences can be discretized and that cognition has a fixed-length reading window. Then, in order for the cognitive concepts to be understood, there must be a certain natural grammar, which is called the form of cognition. Almost all languages use the subject-verb-object triad, and subject-object descriptions are powerful enough to describe all cognitions that can be understood. As a result, there is reason to believe that the fundamental form of cognition is a triple of subject, object, and relation.\par
\begin{figure}
  \centering
  \includegraphics[scale=0.7]{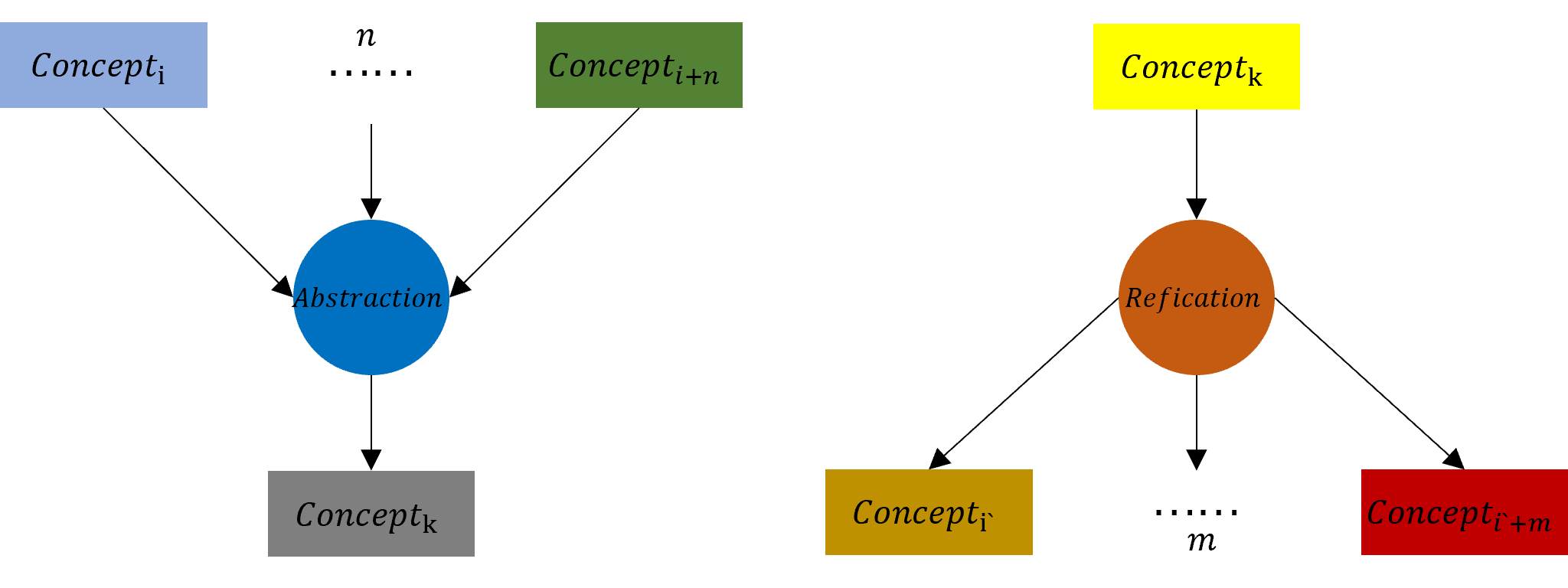}
  \caption{Abstraction and Reification}
\end{figure}
Since the cognitive window size is fixed, understanding complex cognitive information requires abstraction to abstract out a semantic to replace the triplet whole that was previously read, and then understanding of more complex information is formed by understanding the relationship between the abstractions of concepts triple. Second, because abstractions obscure details, it is sometimes necessary to rematerialize abstract concepts in order to comprehend or manipulate them, in which case the previously abstract concepts are re-reduced to basic triples. The cognitive level is extended upward and downward through abstraction and representation. The basic reasoning ability is created by combining the two functions of abstraction and reification.
\begin{figure}
  \centering
  \includegraphics[scale=0.6]{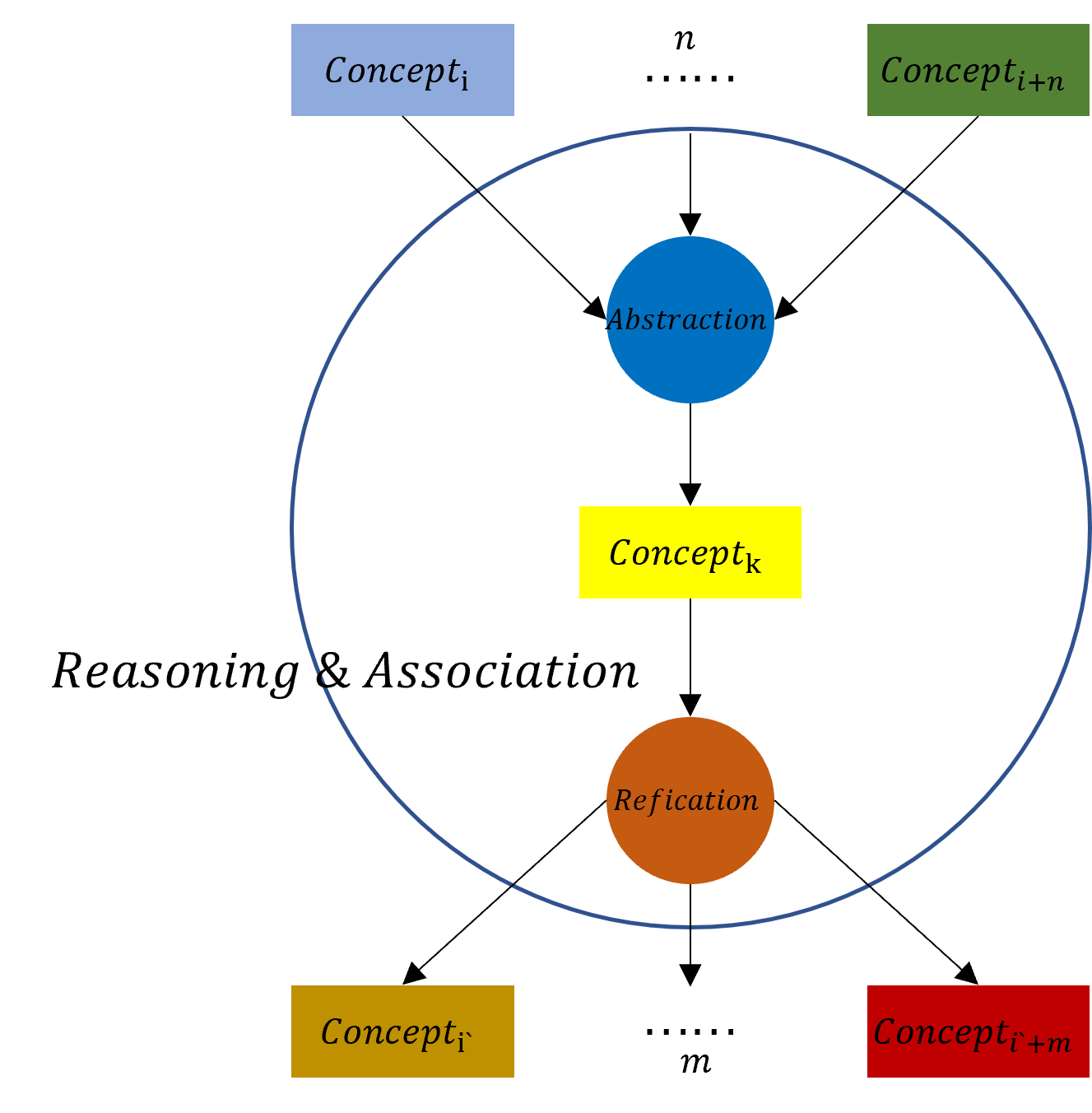}
  \caption{Abstraction and Reification}
\end{figure}

\subsection{Existence of small logic and small tasks in cognition}
\begin{quote}
    \textit{Humans may think so quickly that they ignore or have difficulty capturing the details of their thinking.}
\end{quote}
Can humans directly solve complex problems? The majority of the facts on the subject are on the no side. Adding and subtracting numbers greater than ten is an example of this. Calculate "32495435834+12312314" as an example. Obviously, the human brain cannot directly add or subtract two large numbers. Humans, on the other hand, can quickly calculate the desired answer by adding, subtracting, and carrying bits. Alternatively, ones can enter numbers into a calculator and then manipulate the calculator to get the answer one needs. Then the problem is solved one step at a time, and each step is relatively simple. Can these small tasks, however, be directly experienced? The following example demonstrates how a human's ability to process small tasks too quickly can cause one didn't realize them. \par
\begin{figure}
    \centering
    \includegraphics[scale=0.5]{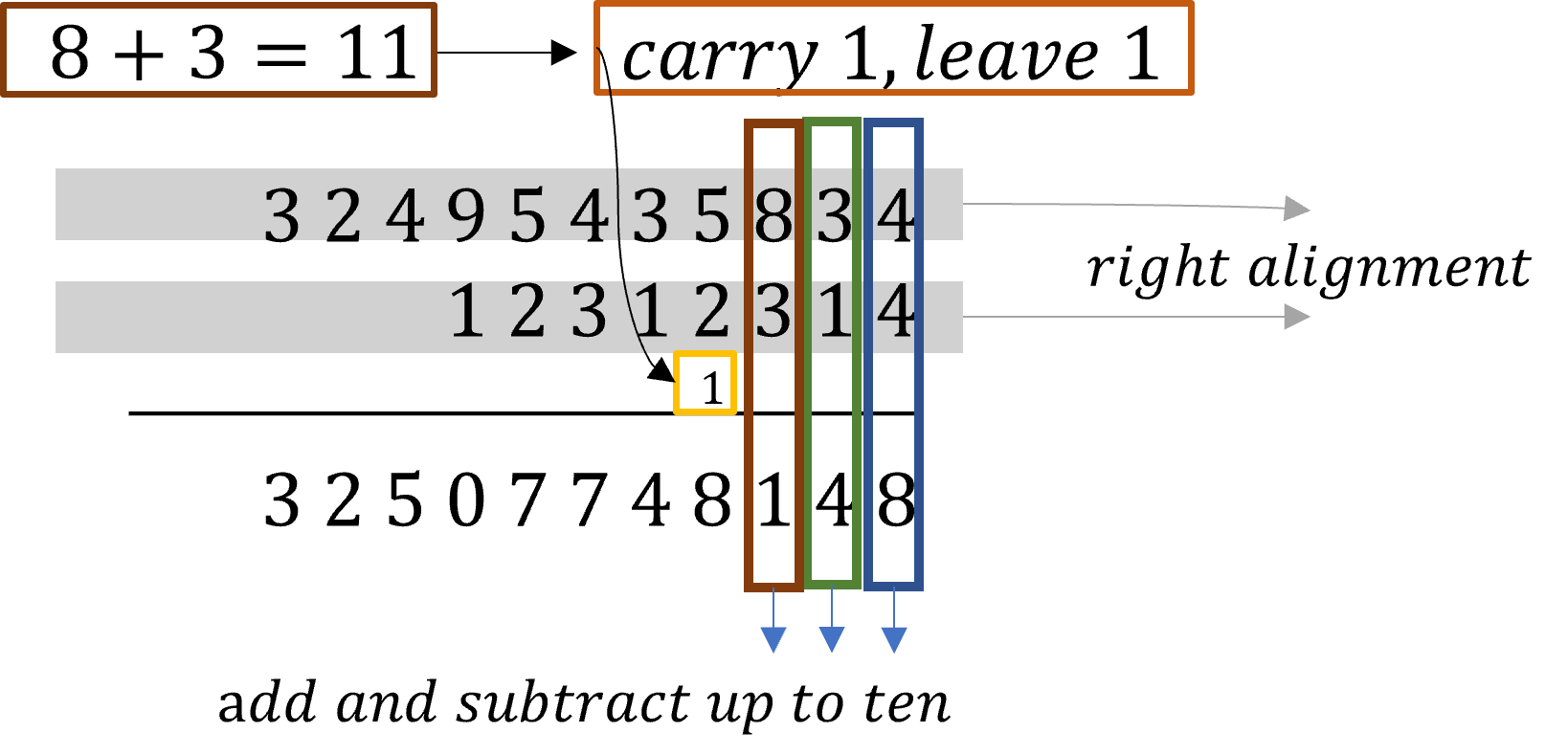}
    \caption{Small tasks in plus}
\end{figure}
Solve or simplify the equation $4 = -x + 5$, for example. This is so simple that even a student who has only studied elementary school maths can find the correct answer $x = 1$ in a matter of seconds. However, when we try to describe this process in natural language, we discover that it is not as simple as performing.
\begin{quote}
1.	Move “– x” from the right side of the “=” to the left side “=”, because of the preference to put the variable at the front of the equation and It is known that “-$\lambda$”, move from the one side of the equation to the other side, the “-” should be removed. Then it becomes “x 4 = $\square$ + 5”.\par
2.	Since there is an empty space “$\square$” before the “+ 5”, therefore, move the “+ 5” to the space and because “+5” is the first part of the right side, then remove the “+” before the “5”. \par
3.	After that, start moving the “4”, because of the preference to move all the numbers to the right side of “=”, and since the number “5” is bigger than “4”, therefore it should be settled behind the “5” and since “4” is moved from the other side of “=”, or stride the “=”, it should be changed to “-4” and then  get “x $\square$ = 5 - 4”. Then remove the “$\square$” and get “x = 5 - 4”.\par
4.	Finally, calculate “5 - 4” and get “1” and use “1” to replace the “5-4” in “x = 5 - 4” and get “x = 1”.
\end{quote}
The formula is simplified with a few small tasks and small logic. The derivation of such small tasks and logic, however, is not based on rules. On the contrary, logic cognition is formalised, whereas derivation is empirical. The benefit of small logic and small tasks is that, while the learning and derivation process is empirical, the outcome is relatively stable due to its small size.

\section{Limitations}

Although Meta Semantic can fully express the logical structure from word to sentence level in the Understanding process, there are some issues in the generation process. " reason, so, x + y + z = 1," for example. When considering the generation of this sentence, we discover that "so" and "x" are closely related in terms of generation timing, but they are logically distant. The topmost structure in this example is (reason, so, result). The words "reason" and "so" were generated at time $t=0,1$, but this meta was finished at time $t=8$. This issue is especially dangerous during the generative process. First, the leaf "x" is generated based on p ("x" | "reason", "so") according to the time sequence to generate. However, from a logical standpoint, the generation of the tree "x + y + z = 1" is based on p ("x + y + z = 1 "|"reason", "so"). If the tree "x+y+z=1" is generated from bottom to up, it conforms to the chronological order of generation but not to the logical relationship from near to far. If the tree is generated from top to bottom, it conforms to the logical relationship from near to far, but there are timing issues; for example, the overall semantics of "x+y+z=1" is derived from "reason" and "so" before the "x". The semantics are then refined so that the order in which they are generated appears in reverse chronological order. "x+y+z=1"->("x+y +z","=", "1") -> (("x+y","+", "z"),"=", "1").

\begin{figure}[h]
    \centering
    \includegraphics[scale=0.6]{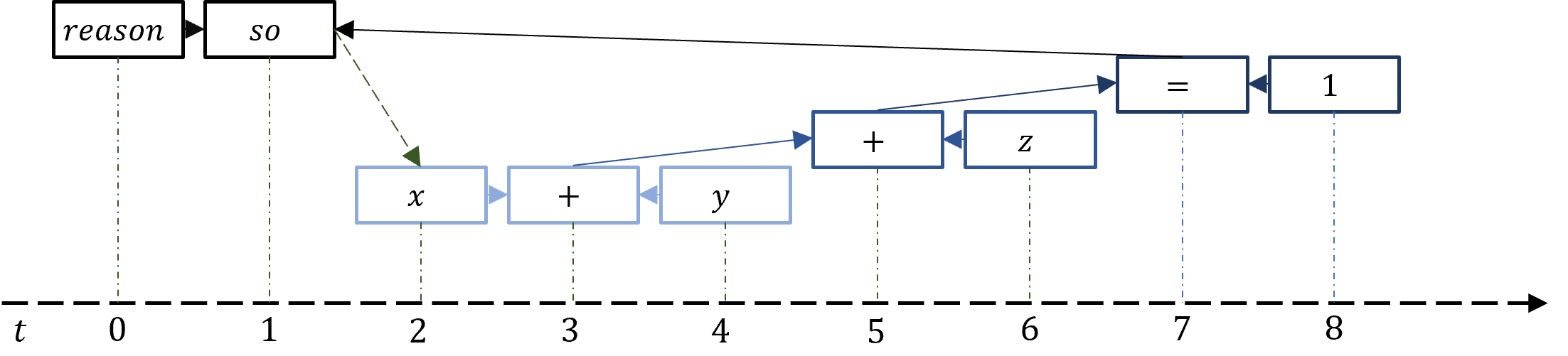}
    \caption{Conflict between logical order and chronological order}
\end{figure}

\section{Conclusion}
We propose two strategies to solve the OOV problem in the statistic-based language model which extends the embedding dictionary to all vocabulary that could be deducted from dictionary words. To strengthen the logical understanding and reasoning ability of the language model, we first propose a semantic model which has the following advantages compared to the traditional language model. Firstly, the Meta semantic can use external rule-based methods to assist reasoning which insure the accuracy of the reasoning that could be done by rules. Secondly, the meta statement is expressed in a formal structure, which indicates the small tasks and small logic behind the language, which will particularly improve empirical reasoning in language. By employing the statistic-based method, we distinguish the model from the conventional rule-based model, resulting in greater flexibility. Furthermore, because all derivation of meta is transparent, their interpretability and observability are more prominent when compared to the traditional black box generation model. Furthermore, the meta-semantic model's dictionary is dynamic. After deducing OOV through rules, OOV can directly obtain its embedding vector through abstraction. In other words, our meta-semantic model broadens the fixed-size dictionary to include all the OOV that can be deduced from the words in the dictionary with their intensional functions, which significantly expands the robustness and reasoning ability of language models.

%Bibliography
\nocite{dietrich2003discrete,riehl2017category,1996generalized}
\bibliographystyle{unsrt}  
\bibliography{references}

\end{document}